\documentclass[10pt,twocolumn,letterpaper]{article}

\usepackage[final]{cvpr}

\usepackage{times}
\usepackage{epsfig}
\usepackage{graphicx}
\usepackage{amsmath,amssymb}
\usepackage{booktabs}
\usepackage{enumitem}
\usepackage{makecell}
\usepackage{adjustbox}
\usepackage{multirow}
\usepackage{caption} % <-- add this
\usepackage[pagebackref,breaklinks,colorlinks]{hyperref}

\begin{document}

% Option A (concise, method-forward)

% Layer-Selective Cross-Attention over CLIP Representation for Generalized AI-Generated Image Detection

% Option B (highlights the insight)

% Which Layers Know? Cross-Attention Aggregation of Multi-Layer CLIP Representations for Deepfake Detection

% Option C (broader framing, stronger claim)

% Beyond the Last Layer: Cross-Attention Layer Aggregation for Generalizable AI-Generated Image Detection

%%%%%%%%% TITLE
\title{Do Transformations Reveal the Truth? Generative Residual Learning for Generalized AI-Generated Image Detection}

% \title{GenRes++: Generative Residual Learning via Neural Tensor Network for Generalizable AI-Generated Image Detection}

\author{
Kutub Uddin\\
University of Michigan\\
Flint, Michigan, USA\\
{\tt\small kutub@umich.edu}
\and
Nusrat Tasnim\\
Korea Aerospace University\\
Goyang, South Korea\\
{\tt\small tasnim.nishu70@kau.kr}
\and
Awais Khan\\
University of Michigan\\
Flint, Michigan, USA\\
{\tt\small mawais@umich.edu}
\and
Mohammad Umar Farooq\\
University of Michigan\\
Flint, Michigan, USA\\
{\tt\small mufarooq@umich.edu}
\and
Khalid Malik\\
University of Michigan\\
Flint, Michigan, USA\\
{\tt\small drmalik@umich.edu}
}

\maketitle
\thispagestyle{empty}

%%%%%%%%% ABSTRACT
\begin{abstract}
The rapid advancement of generative AI has enabled the creation of highly realistic deepfake media, posing significant threats, including misinformation, digital identity theft, fraud, and manipulation of public opinion. AI-generated image (AIGI) detection is reliably challenging due to the diversity of generative methods and the subtle artifacts they leave behind. In this work, we propose GenRes, a novel framework for generative residual learning via a neural tensor network, which models fine-grained relational features between original and transformed samples to enhance generalization. To address scenarios involving multiple generative transformations, we introduce GenRes++, which employs a learnable attention mechanism to aggregate relational features across multiple transformed samples and enables the model to focus on the most informative cues. Both models leverage PE-Core as a feature extractor, providing generalized and semantically rich embeddings that improve cross-domain performance and enable the detection of AIGI generated by unseen methods. Comprehensive experiments on multiple benchmark datasets demonstrate that the proposed GenRes++ approach outperforms existing methods.
\end{abstract}
\vspace{-25pt}
\begin{figure}[!t]
\centering
\begin{tabular}{cccccc}
\includegraphics[width=7cm]{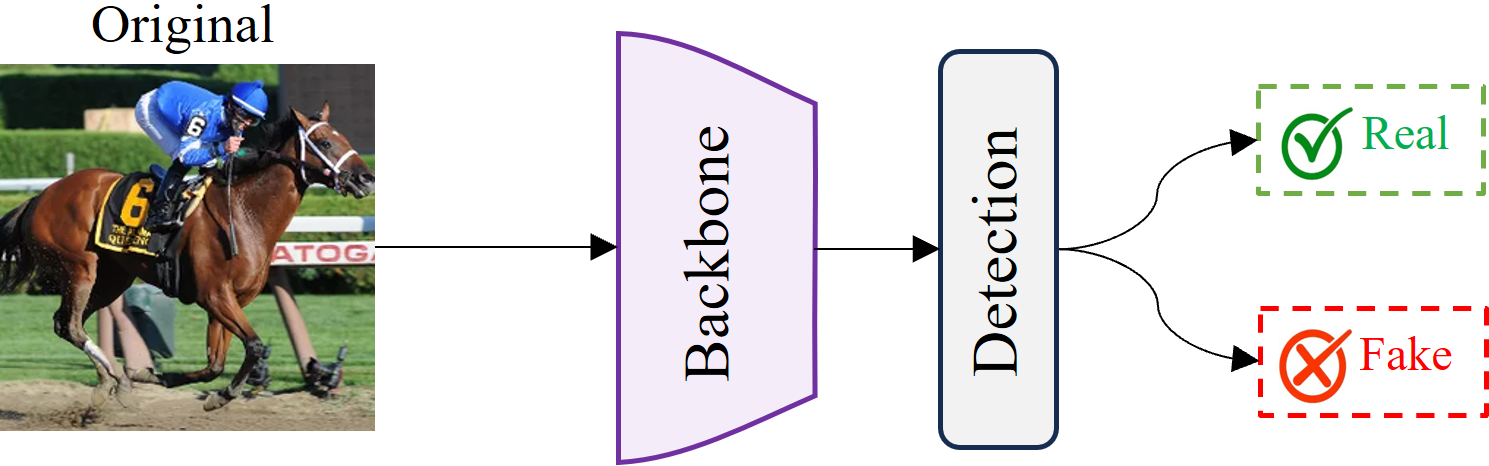}\label{fig:intro_a} \vspace{-5pt}\\ 
(a) \\
\vspace{-5pt}
\includegraphics[width=7cm]{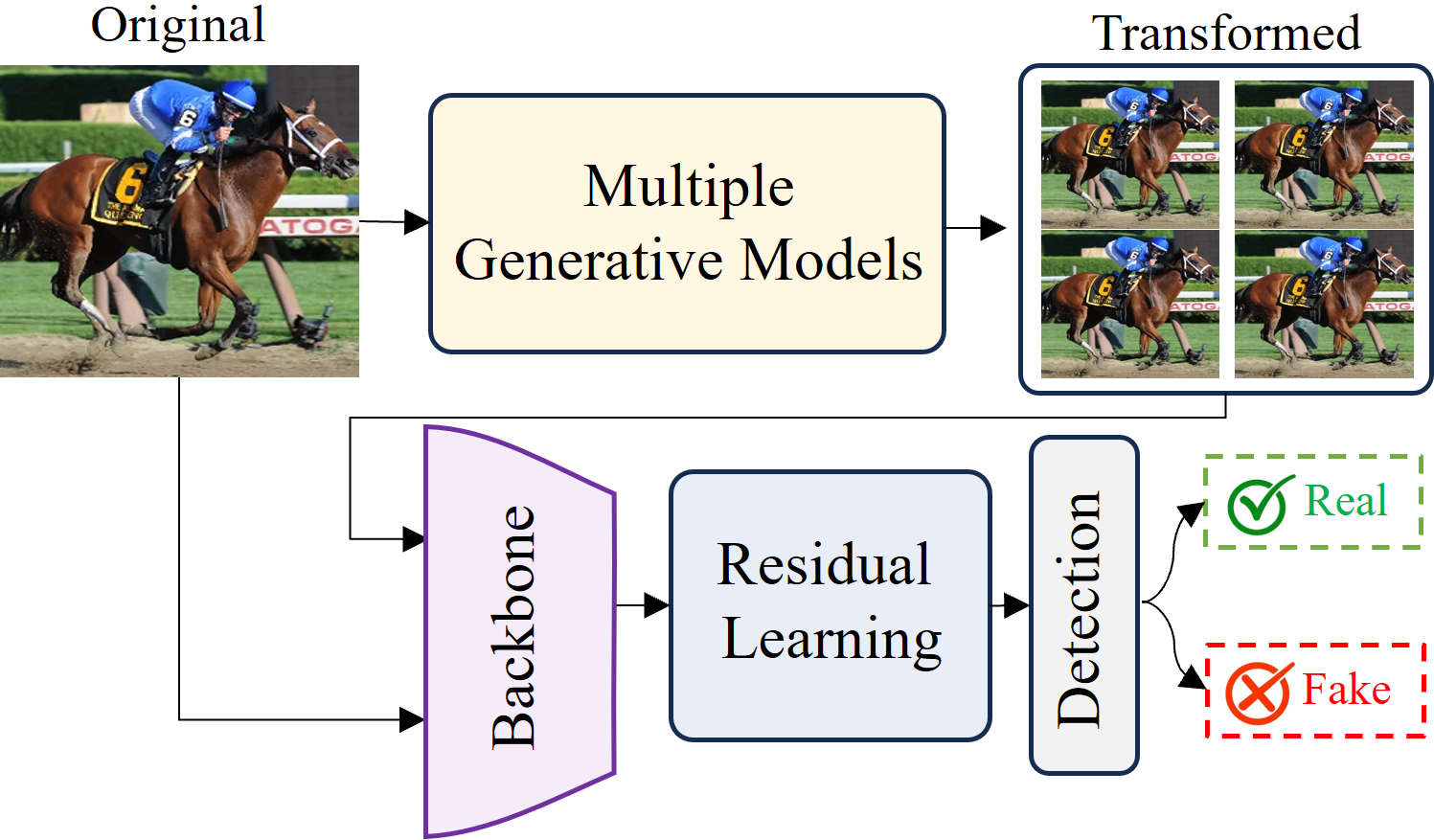} \vspace{-5pt}\\
(b)\\
\end{tabular}
\caption{Comparison between traditional AIGI detection and the proposed \textbf{GenRes++} framework. (a) Traditional approaches encode a single image and perform direct detection, relying on isolated features, thereby limiting their ability to capture subtle generative artifacts and weakening cross-generator generalization. (b) In contrast, our approach generates multiple transformed variants and learns relational residuals between the original and transformed images. By exploiting how generative artifacts interact under transformations, the model captures more discriminative properties, leading to improved generalization.}
  \label{fig:intro}
\vspace{-20pt}
\end{figure}
% \vspace{-10pt}

%%%%%%%%% BODY TEXT
\section{Introduction}\vspace{-5pt}
Advances in generative adversarial networks (GANs)~\cite{karras2017progressive}, diffusion models~\cite{ho2020denoising}, and hybrid~\cite{esser2021taming} synthesis pipelines have made AIGIs increasingly indistinguishable from real photographs, fueling misinformation, identity fraud, and deepfake-related financial losses now reaching hundreds of millions of dollars annually~\cite{esecurityplanet2025,eftsureForecast,tasnim2026comprehensive,uddin2025adversarial}. Detecting such content reliably is therefore an urgent and practically consequential problem. Existing detectors fall into two paradigms: artifact-driven methods exploit generator-specific fingerprints~\cite{frank2020leveraging,durall2020watch,tan2024rethinking, tasnim2026diversity,uddin2025sheild, uddin2023deep}, but degrade sharply on unseen generators, while representation-learning methods project images into forgery-sensitive CLIP spaces~\cite{ojha2023towards,cozzolino2024raising,tan2025c2p, uddin2026face2parts} yet still encode each image independently, neglecting the relational structure induced by generative processing. As a result, neither paradigm achieves generalization across the rapidly expanding diversity of generative models.\\
Real and synthetic images respond differently when processed by a second generative model (restoration, super-resolution, enhancement, or denoising)~\cite{uddin2023robust, uddin2024counter}. Natural images produce outputs broadly consistent with their rich, high-frequency content, whereas AIGIs exhibit characteristic residual discrepancies as the transform interacts with the original generator's statistical biases~\cite{tasnim2025ai,tasnim2026grex,farooq2026trace,khan2026dual, uddin2025advbench}. This occurs because an AIGI already embodies the distributional priors of its source generator. When a second generative model is applied, these embedded biases interact with the inductive assumptions of the transformation network in a non-trivial, detectable way. Crucially, this differential response is a general property of synthetic images. It holds across generator families, whether the source is a GAN~\cite{gandhi2020adversarial, mi2020gan} or a diffusion model~\cite{nichol2021improved}, making it a naturally cross-generator detection property. We term these discrepancies \emph{generative residuals} and build our framework around explicitly modeling them rather than detecting images in isolation.\\
We propose \textbf{GenRes}, which models the relational structure between an original image and its transformation via a neural tensor network (NTN)~\cite{socher2013reasoning} that captures multiplicative cross-feature dependencies beyond simple subtraction or concatenation. This bilinear interaction exposes correlated perturbations across feature dimensions that are characteristic of synthetic provenance yet not tied to any specific generator's artifacts. Its extension, \textbf{GenRes++}, adds a cross-attention aggregation ($CCA$)~\cite{vaswani2017attention} module to pool relational properties from $N$ diverse transforms adaptively, directing the model toward the most informative residuals for each input. Both models use a frozen \textbf{PE-Core ViT}~\cite{bolya2025perception} with LoRA~\cite{hu2021lora} fine-tuning, providing semantically rich embeddings that further strengthen cross-generator generalization at low parameter cost. Comprehensive experiments on the UniversalFakeDetect~\cite{wang2020cnn} benchmark demonstrate that our method outperforms existing approaches on 19 unseen generative models. Our contributions are:\vspace{-5pt}
\begin{itemize}
    \item We propose GenRes, a relational representation learning framework that models discrepancies between original and transformed images to enhance cross-generator generalization in AIGI detection.\vspace{-5pt}
    \item We employ an NTN to capture fine-grained bilinear interactions between feature embeddings of original and transformed samples to explore intrinsic relationships.\vspace{-6pt}
    \item We further extend the framework to GenRes++, incorporating a cross-attention aggregation mechanism to adaptively fuse multiple transformed variants and improve generalization under unseen generative artifacts. \vspace{-20pt}
    \item We conduct comprehensive evaluations on a diverse benchmark comprising 19 unseen generative models, demonstrating strong cross-generator generalization and consistent gains over existing methods.
\end{itemize}
%------------------------------------------------------------------------
\section{Related Work}\vspace{-5pt}
In this section, we provide an overview of AIGI detection methods, including artifact-based, forgery-focused, pre-trained, and cross-generator approaches.
% This section provides a comprehensive review of existing AIGI detection methods spanning artifact-driven approaches, pre-trained representation learning, and cross-generator generalization strategies.
\vspace{-3pt}
\subsection{Artifact-Driven AI-Generated Image Detection}\vspace{-5pt}
Early AIGI detectors~\cite{wang2020cnn,durall2020watch,frank2020leveraging} exploit low-level statistical fingerprints specific to generative pipelines. In particular, Wang~\etal~\cite{wang2020cnn} (CNN-Spot) showed that a ResNet-50 trained on a single GAN with aggressive blur and JPEG augmentation generalizes across GAN architectures, establishing the foundational cross-generator evaluation protocol. Frank~\etal~\cite{frank2020leveraging} and Durall~\etal~\cite{durall2020watch} exploit frequency artifacts in the DCT spectrum, while FreqNet~\cite{tan2024frequency} extends this by learning spectral residuals in the frequency domain. LGrad~\cite{tan2023learning} leverages gradient-based representations to expose upsampling traces, and NPR~\cite{tan2024rethinking} revisits up-convolution artifacts in CNN-based generators to improve cross-generator generalization. Nataraj~\etal~\cite{nataraj2019detecting} exploit co-occurrence patterns across color channels as a complementary low-level cue. While these artifact-driven approaches are effective within the generator family seen during training, they do not transfer reliably across families: diffusion models synthesize images through an iterative denoising process that does not share the upsampling or spectral characteristics of GANs, causing such detectors to degrade significantly on diffusion-generated content~\cite{corvi2023detection}.
% \vspace{-4pt}
\subsection{Forgery-Focused and Attention-based Methods}\vspace{-5pt}
Recent work emphasizes learning discriminative representations to improve cross-generator generalization. For instance, PatchForensics~\cite{chai2020makes} proposes a patch-level classifier that limits the receptive field to expose local forgery artifacts, improving generalization across manipulation types. In contrast, F3Net~\cite{qian2020thinking} decomposes the frequency spectrum into distinct bands and mines forgery-sensitive frequency components via cross-band attention. FatFormer~\cite{liu2024forgery} introduces a forgery-aware adaptive transformer that recalibrates attention toward synthesis traces in both local and global feature spaces. Recently, RINE~\cite{koutlis2024leveraging} provides strong evidence that intermediate encoder-block representations carry complementary forgery cues at different levels of abstraction, motivating the use of multi-layer features for generalizable detection.
\vspace{-6pt}
\subsection{Pre-Trained Representation Learning}\vspace{-5pt}
A more generalizable paradigm projects images into forgery-sensitive embedding spaces derived from large-scale pre-trained models. UniFD~\cite{ojha2023towards} demonstrated that frozen CLIP features implicitly encode artifact-indicative information that transfers across generators without any retraining. Cozzolino~\etal~\cite{cozzolino2024raising} conducted a systematic analysis of CLIP's discriminative power and proposed targeted fine-tuning strategies to raise the detection bar against modern synthesis methods. C2P-CLIP~\cite{tan2025c2p} injects category-common prompts to sharpen the decision boundary across diverse generators, while ForgeLens~\cite{chen2025forgelensdataefficientforgeryfocus} proposes a data-efficient strategy that focuses attention on forgery-relevant regions. EFFORT~\cite{yan2024effort} applies parameter-efficient fine-tuning to a vision transformer backbone, demonstrating that LoRA-style adaptation preserves pre-trained generalization while enabling task-specific refinement. DIRE~\cite{wang2023dire} takes a complementary residual-based approach, reconstructing images through a diffusion model and using the pixel-level reconstruction error for AIGI detection.
\vspace{-6pt}
\subsection{Cross-Generator Generalization}\vspace{-5pt}
Maintaining high detection accuracy on generative models never seen during training is the primary practical challenge for deployed AIGI detectors. The community has converged on training on a single generator (typically ProGAN~\cite{karras2017progressive}) and evaluating on a broad suite of unseen generators as the standard protocol~\cite{wang2020cnn,ojha2023towards}, spanning 19 diverse unseen generative models, including GANs, diffusion, and low-level vision and perceptual models. Rajan~\etal~\cite{rajan2024aligned} show that aligning training and test distributions substantially improves detection of latent diffusion outputs, while~\cite{rajan2025stay} argues for discarding real-image features entirely to reduce domain bias. 
% Despite this progress, most methods treat detection as an independent encoding problem, extracting features from each image in isolation. This creates a structural limitation: features learned from a single training generator inherit its generative artifacts, leading to poor generalization. Our work addresses this by leveraging relational generative residuals, the differential response of real and synthetic images under secondary generative processing, a signal that generalizes across generator families without requiring exposure to the target generator during training.
Despite recent progress, most methods treat detection as an independent feature extraction problem on single images, which often leads to overfitting to generator-specific artifacts and weak generalization. We address this by modeling relational generative residuals, the differences between real and synthetic images under a secondary transformation which provides a more general signal that transfers across generator families without requiring access to target generators during training.

\begin{figure*}[t]
\centering
   \includegraphics[width=1\linewidth]{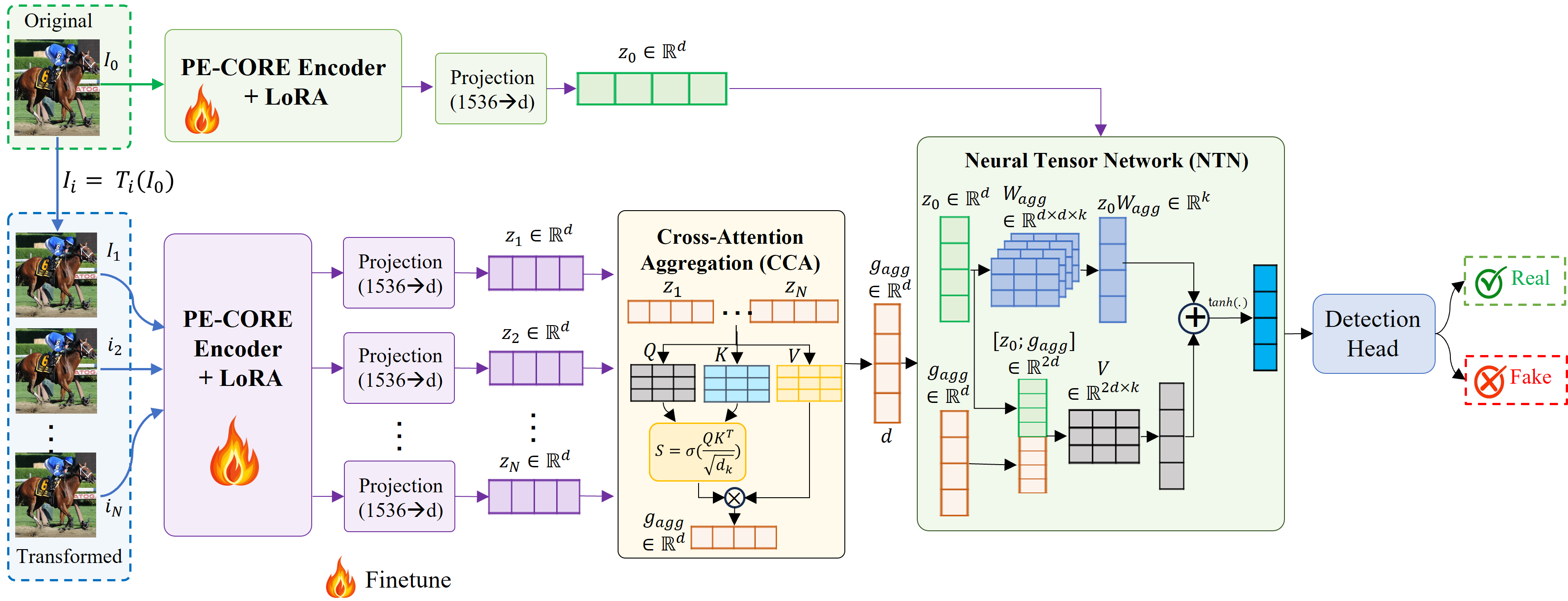}
   \caption{Overview of the proposed $GenRes++$ framework. The original image $I_0$ and its generative transformed variants $\{I_i\}_{i=1}^{N}$ are encoded using a shared PE-CORE encoder with LoRA adaptation, followed by a projection into a compact feature space. The transformed features are aggregated via a $CCA$ module to produce a unified representation $\mathbf{g}_{\text{agg}}$. A NTN then models the bilinear relational interaction between the original embedding $\mathbf{z}_0$ and $\mathbf{g}_{\text{agg}}$, capturing fine-grained generative residuals. The resulting representation is fed to a detection head to classify the input as real or AIGI, respectively.}
\label{fig:arch}
\vspace{-8pt}
\end{figure*}
\vspace{-6pt}
\section{Proposed Method}\vspace{-5pt}
This section presents the motivation, architecture, and training objective of GenRes and GenRes++.
\vspace{-3pt}
\subsection{Motivation and Problem Formulation}\vspace{-5pt}
Existing AIGI detectors encode each image independently, extracting forgery cues in isolation~\cite{ojha2023towards,wang2020cnn,tan2025c2p}. This is a structural limitation: a detector trained on one generator learns that generator's specific artifact signature, which does not transfer to unseen generators~\cite{corvi2023detection}. We address this by exploiting a complementary signal, the differential response of real versus AIGI images under secondary generative processing. When an image is passed through a second generative model~\cite{wang2021gfpgan,wang2021realesrgan,zhang2018ffdnet}, such as restoration, super-resolution, or denoising, the output depends on the statistical structure of the input. Real images, carrying rich, naturally distributed content, produce outputs broadly consistent with the input. AI-generated images, however, already embody the distributional biases of their source generator~\cite{karras2017progressive,ho2020denoising}; when subjected to a second generative model, these biases interact with the inductive priors of the transformation network in a detectable, non-trivial way~\cite{wang2023dire}. We term this discrepancy a generative residual. Crucially, this differential response is a general property of synthetic images; it holds across GAN and diffusion families like~\cite{frank2020leveraging,corvi2023detection}, making it a naturally cross-generator detection signal.\\
Formally, let $I_0$ denote an input image and $\{T_i\}_{i=1}^{N}$ a collection of $N$ generative transforms, each producing $I_i = T_i(I_0)$. The detection task exploits the pairwise relationships $\{(I_0, I_i)\}_{i=1}^{N}$ as evidence. A key design requirement is that the relational feature between $I_0$ and $I_i$ should capture multiplicative cross-dimensional dependencies rather than merely element-wise differences~\cite{socher2013reasoning}, since generative residuals manifest as correlated perturbations across feature dimensions. This motivates the use of a Neural Tensor Network~\cite{socher2013reasoning} as the relational interaction module.\vspace{-5pt}
\subsection{Architecture Overview}\vspace{-5pt}
The GenRes++ architecture consists of four main components: (1) a shared PE-Core vision encoder with LoRA fine-tuning, (2) a bottleneck projection layer, (3) a $CCA$ module, and (4) an NTN followed by a binary classification head. The full architecture is illustrated in Figure~\ref{fig:arch}.\vspace{-12pt}
\subsubsection{Shared Vision Encoder with LoRA}\vspace{-4pt}
All images, both the original $I_0$ and the $N$ transformed variants $\{I_i\}_{i=1}^{N}$, are encoded by a single shared PE-Core-G14-448 encoder~\cite{bolya2025perception}, a vision transformer model with embedding dimension $d_\text{enc} = 1536$. The encoder is kept frozen during training to preserve the rich, semantically structured representation learned during pre-training. Parameter-efficient adaptation is achieved via LoRA~\cite{hu2021lora} applied to the query (Q), key (K), and value (V) projection matrices of every self-attention layer. LoRA introduces lightweight rank-$r$ update matrices $A$ and $B$ such that the adapted weight is $W' = W + \frac{\alpha}{r} AB$, where only $A$ (down-projection) and $B$ (up-projection) are trained. We empirically set $r = 6$ and $\alpha = 8$ throughout, keeping the trainable parameter low while enabling task-specific feature adaptation.\\
Following encoding, each feature vector is $\ell_2$-normalized and projected from $d_\text{enc} = 1536$ to a compact embedding dimension $d$ via a learned linear projection layer initialized with small normal noise. The projected embedding for the original image is denoted $\mathbf{z}_0 \in \mathbb{R}^d$, and the projected embeddings for the transformed images are $\mathbf{z}_1, \ldots, \mathbf{z}_N \in \mathbb{R}^d$.\vspace{-12pt}
\subsubsection{Cross-Attention Aggregation}\vspace{-5pt}
In GenRes++, the $N$ transformed embeddings $\{\mathbf{z}_1, \ldots, \mathbf{z}_N\}$ are aggregated into a single representative vector $\mathbf{g}_\text{agg} \in \mathbb{R}^d$ using the $CCA$ module. The $CCA$ treats the original embedding $\mathbf{z}_0$ as a query and the $N$ transformed embeddings as keys and values. Let $Z = [\mathbf{z}_1, \ldots, \mathbf{z}_N] \in \mathbb{R}^{N \times d} (d=256)$. The attention logits are computed as:
\begin{equation}
    S = \sigma\!\left(\frac{Q K^\top}{\sqrt{d_k}}\right),
\end{equation}
where $Q = \mathbf{z}_0 W_Q$, $K = Z W_K$, $V = Z W_V$, $W_Q, W_K, W_V$ are learned projection matrices, $d_k = d/4$ is the per-head dimension across 4 attention head, and $\sigma$ denotes softmax over the $N$ transform dimension. The aggregated representation is:\vspace{-5pt}
\begin{equation}
    \mathbf{g}_\text{agg} = S V \in \mathbb{R}^d.
\end{equation}
This mechanism allows the model to dynamically weight each transformed variant according to how informative its residual is relative to the specific query image, rather than treating all transforms as equally relevant.\vspace{-10pt}
\subsubsection{Neural Tensor Network}\vspace{-5pt}
The relational interaction between $\mathbf{z}_0$ and $\mathbf{g}_\text{agg}$ is modeled by an NTN. Given $\mathbf{z}_0, \mathbf{g}_\text{agg} \in \mathbb{R}^d$, the NTN computes:
\begin{equation}
    \mathbf{f} = \tanh\!\left(\mathbf{z}_0^\top \mathbf{W} \, \mathbf{g}_\text{agg} + [\mathbf{z}_0;\, \mathbf{g}_\text{agg}]^\top V + \mathbf{b}\right),
\end{equation}
where $\mathbf{W} \in \mathbb{R}^{d \times d \times k}$ is a third-order weight tensor that captures pairwise interactions across all feature dimension pairs, $V \in \mathbb{R}^{2d \times k}$ is a conventional linear weight matrix applied to the concatenation $[\mathbf{z}_0;\, \mathbf{g}_\text{agg}]$, $\mathbf{b} \in \mathbb{R}^k$ is a bias vector, $k=128$ is the NTN output dimension, and $\tanh$ is applied element-wise. The resulting vector $\mathbf{f} \in \mathbb{R}^k$ encodes the full relational structure between the original and aggregated transformed representations. The bilinear term $\mathbf{z}_0^\top \mathbf{W} \, \mathbf{g}_\text{agg}$ is the key distinction from simpler fusion strategies. While element-wise subtraction captures first-order additive differences and concatenation with a linear layer captures weighted sums, the tensor product captures multiplicative cross-feature dependencies: the response of feature dimension $i$ in $\mathbf{z}_0$ modulated by feature dimension $j$ in $\mathbf{g}_\text{agg}$, for all pairs $(i,j)$. This is precisely the structure of generative residuals, where artifacts in one feature subspace interact multiplicatively with representations in another subspace. The NTN output $\mathbf{f} \in \mathbb{R}^k$ is passed through a linear layer to produce a scalar logit: \vspace{-5pt}
\begin{equation}
    \ell = \mathbf{w}_\text{head}^\top \mathbf{f} + b_\text{head},
\end{equation}
where $\mathbf{w}_\text{head} \in \mathbb{R}^k$ and $b_\text{head} \in \mathbb{R}$ are learned parameters. At inference time, $\ell$ is passed through a sigmoid to yield a detection probability to be classified as real and AIGI. \vspace{-10pt}
\subsubsection{Training Objectives}\vspace{-5pt}
Both GenRes and GenRes++ are trained with a binary cross-entropy ($BCE$) with logits loss:
\begin{equation}
    \mathcal{L} = -\left[y \log \sigma(\ell) + (1 - y) \log (1 - \sigma(\ell))\right],
\end{equation}
where $y \in \{0, 1\}$ is the ground-truth label ($0$ = real, $1$ = fake) and $\sigma$ denotes the sigmoid function.\\
\vspace{-7pt}

\begin{figure*}[!t]
\centering
   \includegraphics[width=0.9\linewidth]{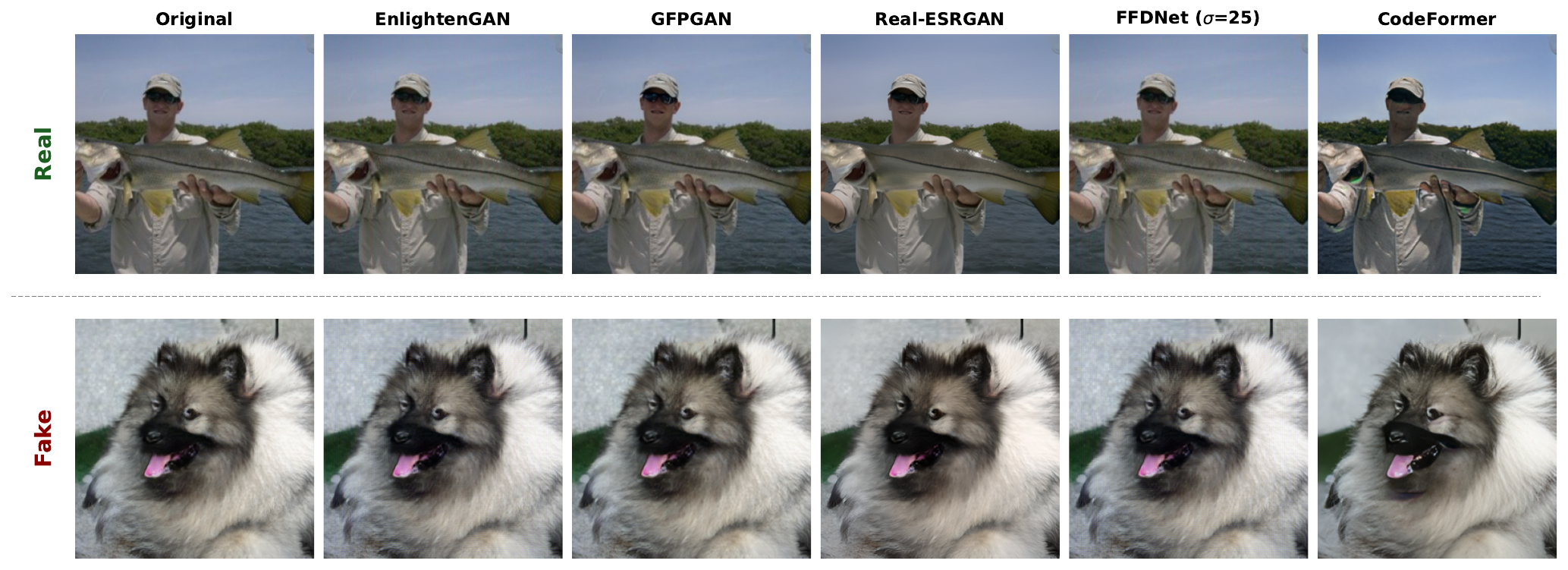}
   \caption{Qualitative comparison of five generative transforms applied to one real and one fake image (BigGAN). Average PSNR/SSIM between the original and each transform: 
EnlightenGAN (32/0.95)~\cite{jiang2021enlightengan}, GFPGAN (44.5/1.00), Real-ESRGAN (34.3/0.96), 
FFDNet (34.8/0.88), CodeFormer (25.4/0.86) for the real and fake samples. Lower PSNR reflects 
stronger modification.}\vspace{-6pt}
% , confirming that the transforms span a wide range of perturbations strengths.}
\label{fig:trans}
\end{figure*}

\begin{table*}[!t]
\centering
\caption{Detection ACC (\%) of different methods across various generative model categories on UniversalFakeDetect~\cite{wang2020cnn}. GenRes++ is trained on ProGAN~\cite{karras2017progressive} 4 object categories and tested on 19 unseen generative models.}
\vspace{5pt}
\resizebox{\textwidth}{!}{
\begin{tabular}{c *{7}{c} *{2}{c} *{2}{c} c *{3}{c} *{3}{c} c c}
\toprule
\midrule
  & \multicolumn{7}{c}{\textbf{GAN Family}}
  & \multicolumn{5}{c}{\textbf{Other Family}}
  & \multicolumn{8}{c}{\textbf{Diffusion Family}} \\
\cmidrule(lr){2-8}
\cmidrule(lr){9-13}
\cmidrule(lr){14-21}
  \rotatebox{90}{Method}
  & \rotatebox{90}{ProGAN}
  & \rotatebox{90}{CycleGAN}
  & \rotatebox{90}{BigGAN}
  & \rotatebox{90}{StyleGAN}
  & \rotatebox{90}{GauGAN}
  & \rotatebox{90}{StarGAN}
  & \rotatebox{90}{Deepfakes}
  & \rotatebox{90}{SITD}
  & \rotatebox{90}{SAN}
  & \rotatebox{90}{CRN}
  & \rotatebox{90}{IMLE}
  & \rotatebox{90}{Guided}
  & \rotatebox{90}{LDM 200}
  & \rotatebox{90}{\makecell{LDM 200\\ CFG}}
  & \rotatebox{90}{LDM 100}
  & \rotatebox{90}{\makecell{Glide 100\\27}}
  & \rotatebox{90}{\makecell{Glide 50\\27}}
  & \rotatebox{90}{\makecell{Glide 100\\10}}
  & \rotatebox{90}{DALL-E}
  & \rotatebox{90}{mACC} \\
\midrule
\midrule
PatchFor~\cite{chai2020makes}  & 75.0 & 69.0 & 68.5 & 79.2 & 64.2 & 63.9 & 75.5 & 75.1 & 75.3 & 72.3 & 55.3 & 67.4 & 76.5 & 76.1 & 75.8 & 74.8 & 73.3 & 68.5 & 67.9 & 71.2 \\
F3Net~\cite{qian2020thinking}    & 99.4 & 76.4 & 65.3 & 92.6 & 58.1 & 100.0 & 63.5 & 54.2 & 47.3 & 51.5 & 51.5 & 96.2 & 68.2 & 75.4 & 68.8 & 81.7 & 83.3 & 83.1 & 66.3 & 71.3 \\
FreqNet~\cite{tan2024frequency}  & 97.9 & 95.8 & 90.5 & 97.6 & 90.2 & 93.4 & 97.4 & 88.9 & 59.0 & 71.9 & 67.4 & 86.7 & 84.6 & 99.6 & 65.6 & 85.7 & 97.4 & 88.2 & 59.1 & 85.1 \\
CNN-Spot~\cite{wang2020cnn} & 100.0 & 85.2 & 70.2 & 85.7 & 79.0 & 91.7 & 53.5 & 66.7 & 48.7 & 86.3 & 86.3 & 60.1 & 54.0 & 55.0 & 54.1 & 60.8 & 63.8 & 65.7 & 55.6 & 69.6 \\
LGrad~\cite{tan2023learning}    & 99.8 & 85.4 & 82.9 & 94.8 & 72.5 & 99.6 & 58.0 & 62.5 & 50.0 & 50.7 & 50.8 & 77.5 & 94.2 & 95.9 & 94.8 & 87.4 & 90.7 & 89.6 & 88.4 & 80.3 \\
NPR~\cite{tan2024rethinking}      & 99.8 & 95.0 & 87.6 & 96.2 & 86.6 & 99.8 & 76.9 & 66.9 & 98.6 & 50.0 & 50.0 & 84.6 & 97.7 & 98.0 & 98.2 & 96.3 & 97.2 & 97.4 & 87.2 & 87.6 \\
UniFD~\cite{ojha2023towards}    & 100.0 & 98.5 & 94.5 & 82.0 & 99.5 & 97.0 & 66.6 & 63.0 & 57.5 & 59.5 & 72.0 & 70.0 & 94.2 & 73.8 & 94.4 & 79.1 & 79.9 & 78.1 & 86.8 & 81.4 \\
FatFormer~\cite{liu2024forgery}& 99.9 & 99.3 & 99.5 & 97.2 & 99.4 & 99.8 & 53.2 & 81.1 & 68.0 & 69.5 & 69.5 & 76.0 & 98.6 & 94.9 & 98.7 & 94.4 & 94.7 & 94.2 & 98.8 & 90.9 \\
RINE~\cite{koutlis2024leveraging}     & 100.0 & 99.3 & 99.6 & 88.9 & 99.8 & 99.5 & 80.6 & 90.6 & 68.3 & 89.2 & 90.6 & 76.1 & 98.3 & 88.2 & 98.6 & 88.9 & 92.6 & 90.7 & 95.0 & 91.3 \\
C2P-CLIP~\cite{tan2025c2p} & 100.0 & 97.3 & 99.1 & 96.4 & 99.2 & 99.6 & 93.8 & 95.6 & 64.4 & 93.3 & 93.3 & 69.1 & 99.3 & 97.3 & 99.3 & 95.3 & 95.3 & 96.1 & 98.6 & 93.8 \\
FreLens~\cite{chen2025forgelensdataefficientforgeryfocus} & 100.0 & 99.2 & 97.7 & 96.6 & 98.8 & 95.2 & 89.0 & 85.8 & 93.8 & 97.2 & 97.6 & 73.3 & 98.7 & 97.0 & 98.9 & 96.1 & 96.2 & 95.4 & 98.3 & 95.0 \\
\midrule
\textbf{GenRes}
& 100.0 & 99.2 & 99.8 & 98.4 & 97.5 & 98.7 & 85.8
& 78.4 & 79.9 & 89.5 & 88.7 & 71.6
& 88.1 & 98.5 & 94.8 & 99.5
& 96.4 & 97.9 & 96.6
& 92.6 \\
\textbf{GenRes++}
& 99.9 & 99.4 & 99.6 & 99.6 & 99.9 & 99.0 & 96.6
& 81.1 & 84.8 & 99.4 & 99.5 & 76.3
& 88.9 & 99.7 & 98.4 & 99.6
& 98.6 & 98.6 & 99.1
& \textbf{95.7} \\
\bottomrule
\end{tabular}%
}\vspace{-6pt}
\label{tab:detection_acc}
\vspace{-5pt}
\end{table*}
\vspace{-5pt}
\section{Results}
\vspace{-5pt}
This section presents the experimental setup, comprehensive evaluation, and extensive ablation studies.
\vspace{-5pt}
\subsection{Experimental Setups}
\vspace{-5pt}
Optimization uses AdamW~\cite{zhou2024towards} with learning rate $4 \times 10^{-4}$, weight decay $0.05$, and betas $(0.9, 0.999)$. A multiplicative learning rate schedule decays the rate by a factor of $0.9$ every 10 epochs. Training employs automatic mixed precision with a gradient scaler to reduce memory consumption and accelerate throughput. Only the LoRA adapters~\cite{hu2021lora}, the bottleneck projection, the CCA~\cite{vaswani2017attention}, the NTN, and the classification head are updated; the frozen backbone is excluded from the optimizer. The training data consists of real and AIGIs generated by ProGAN~\cite{karras2017progressive}. We evaluate the proposed GenRes++ on the UniversalFakeDetect dataset~\cite{wang2020cnn}, which comprises 19 unseen generators spanning three families: GANs, including ProGAN~\cite{karras2017progressive}, CycleGAN~\cite{zhu2017unpaired}, 
BigGAN~\cite{brock2018large}, StyleGAN~\cite{karras2019style}, GauGAN~\cite{park2019semantic}, and StarGAN~\cite{choi2018stargan}; other generators, including 
Deepfakes~\cite{rossler2019faceforensics}, SITD~\cite{chen2018learning}, SAN~\cite{dai2019second}, CRN~\cite{chen2017photographic}, and IMLE~\cite{li2019diverse}; and diffusion models, including 
Guided~\cite{dhariwal2021diffusion}, LDM~\cite{rombach2022high}, Glide~\cite{nichol2021glide}, and DALL-E~\cite{ramesh2021zero}.\\
For each image, $N=5$ transformed variants are produced by EnlightenGAN (enhancement)~\cite{jiang2021enlightengan}, GFPGAN (restoration)~\cite{wang2021gfpgan}, Real-ESRGAN (super-resolution)~\cite{wang2021realesrgan}, FFDNet (denoising, $\sigma=25$)~\cite{zhang2018ffdnet}, and CodeFormer (restoration)~\cite{zhou2022codeformer, uddin2022double}. The same five transforms are applied identically to both real and AIGIs, ensuring that the model learns to detect the differential response rather than the transforms themselves. Random horizontal flipping is applied during training; inference uses center-cropped, fixed-size images without augmentation.

\begin{table*}[!t]
\centering
\caption{Detection AP (\%) of different methods across various generative model categories on UniversalFakeDetect~\cite{wang2020cnn}. GenRes++ is trained on ProGAN~\cite{karras2017progressive} 4 object categories and tested on 19 unseen generative models.}
\vspace{4pt}
\resizebox{\textwidth}{!}{
\begin{tabular}{c *{7}{c} *{2}{c} *{2}{c} c *{3}{c} *{3}{c} c c}
\toprule
\midrule
  & \multicolumn{7}{c}{\textbf{GAN Family}}
  & \multicolumn{5}{c}{\textbf{Other Family}}
  & \multicolumn{8}{c}{\textbf{Diffusion Family}} \\
\cmidrule(lr){2-8}
\cmidrule(lr){9-13}
\cmidrule(lr){14-21}
  \rotatebox{90}{Method}
  & \rotatebox{90}{ProGAN}
  & \rotatebox{90}{CycleGAN}
  & \rotatebox{90}{BigGAN}
  & \rotatebox{90}{StyleGAN}
  & \rotatebox{90}{GauGAN}
  & \rotatebox{90}{StarGAN}
  & \rotatebox{90}{Deepfakes}
  & \rotatebox{90}{SITD}
  & \rotatebox{90}{SAN}
  & \rotatebox{90}{CRN}
  & \rotatebox{90}{IMLE}
  & \rotatebox{90}{Guided}
  & \rotatebox{90}{LDM 200}
  & \rotatebox{90}{\makecell{LDM 200\\ CFG}}
  & \rotatebox{90}{LDM 100}
  & \rotatebox{90}{\makecell{Glide 100\\27}}
  & \rotatebox{90}{\makecell{Glide 50\\27}}
  & \rotatebox{90}{\makecell{Glide 100\\10}}
  & \rotatebox{90}{DALL-E}
  & \rotatebox{90}{mAP} \\
\midrule
PatchFor~\cite{chai2020makes}  & 80.9 & 72.8 & 71.7 & 85.8 & 66.0 & 69.3 & 76.6 & 76.2 & 76.3 & 74.5 & 68.5 & 75.0 & 87.1 & 86.7 & 86.4 & 85.4 & 83.7 & 78.4 & 75.7 & 77.7 \\
F3Net~\cite{qian2020thinking}    & 100.0 & 84.3 & 69.9 & 99.7 & 56.7 & 100.0 & 78.8 & 52.9 & 46.7 & 63.4 & 64.4 & 70.5 & 73.8 & 81.7 & 74.6 & 89.8 & 91.0 & 90.9 & 71.8 & 76.9 \\
FreqNet~\cite{tan2024frequency}  & 99.9 & 96.3 & 96.1 & 99.9 & 99.7 & 98.6 & 99.9 & 94.4 & 74.6 & 80.1 & 75.7 & 96.3 & 96.1 & 100.0 & 62.3 & 99.8 & 99.8 & 96.4 & 77.8 & 91.9 \\
CNN-Spot~\cite{wang2020cnn} & 100.0 & 93.5 & 84.5 & 99.5 & 89.5 & 98.2 & 89.0 & 73.8 & 59.5 & 98.2 & 98.4 & 73.7 & 70.6 & 71.0 & 70.5 & 80.7 & 84.9 & 82.1 & 70.6 & 83.6 \\
LGrad~\cite{tan2023learning}    & 100.0 & 93.9 & 90.0 & 79.4 & 36.4 & 100.0 & 67.9 & 59.4 & 42.1 & 63.5 & 69.6 & 87.1 & 99.0 & 99.2 & 99.6 & 93.2 & 95.1 & 94.9 & 97.2 & 86.4 \\
NPR~\cite{tan2024rethinking}      & 100.0 & 99.5 & 94.5 & 99.9 & 88.8 & 100.0 & 84.4 & 97.9 & 99.9 & 50.2 & 50.2 & 98.3 & 99.9 & 99.9 & 99.9 & 99.9 & 99.9 & 99.9 & 99.3 & 92.8 \\
UniFD~\cite{ojha2023towards}    & 100.0 & 98.1 & 94.5 & 86.7 & 99.3 & 99.5 & 91.7 & 78.5 & 67.5 & 83.1 & 91.1 & 79.2 & 95.8 & 79.8 & 95.9 & 93.9 & 95.1 & 94.6 & 88.5 & 90.1 \\
FatFormer~\cite{liu2024forgery}& 100.0 & 100.0 & 99.9 & 99.8 & 100.0 & 100.0 & 97.9 & 97.9 & 81.2 & 99.8 & 99.9 & 91.9 & 99.8 & 99.1 & 99.9 & 99.1 & 99.4 & 99.2 & 99.8 & 98.2 \\
RINE~\cite{koutlis2024leveraging}     & 100.0 & 99.9 & 99.9 & 40.0 & 100.0 & 100.0 & 97.9 & 97.2 & 94.9 & 97.3 & 99.7 & 96.4 & 99.8 & 98.3 & 99.9 & 98.8 & 98.8 & 98.9 & 99.3 & 98.8 \\
C2P-CLIP~\cite{tan2025c2p} & 100.0 & 100.0 & 99.9 & 99.5 & 100.0 & 100.0 & 98.6 & 98.9 & 84.6 & 99.9 & 99.9 & 94.1 & 99.9 & 99.8 & 99.9 & 99.7 & 99.8 & 99.8 & 99.9 & 98.7 \\
FreLens~\cite{chen2025forgelensdataefficientforgeryfocus} & 100.0 & 100.0 & 99.8 & 99.8 & 99.9 & 100.0 & 95.4 & 94.2 & 98.7 & 99.9 & 99.9 & 92.9 & 99.9 & 99.5 & 99.9 & 99.1 & 99.5 & 99.3 & 99.8 & 98.8 \\
\midrule
\textbf{GenRes}
& 100.0 & 99.8 & 100.0 & 99.8 & 98.8 & 99.5 & 94.2
& 89.6 & 88.9 & 97.7 & 96.2 & 86.8
& 97.4 & 100.0 & 98.2 & 100.0
& 99.1 & 98.4 & 98.6
& 97.0 \\
\textbf{GenRes++} & 100.0 & 99.9 & 100.0 & 100.0 & 100.0 & 99.9 & 99.0
& 95.9 & 99.2 & 99.9 & 100.0 & 92.5
& 98.5 & 99.9 & 99.6 & 99.9
& 99.8 & 99.8 & 99.8
& \textbf{99.1} \\
\midrule
\bottomrule
\end{tabular}%
}\vspace{-15pt}
\label{tab:detection_ap}
\end{table*}

\begin{table}[t]
\centering
\caption{Computational Cost of Transforms and Detection.}
\label{tab:compute_cost}
\begin{adjustbox}{width=0.8\linewidth}
\begin{tabular}{lccc}
\toprule\midrule
Transform & Params (M) & GFLOPs & Time (ms) \\
\midrule
EnlightenGAN  & 8.6   & 16.5  & 5.2  \\
GFPGAN        & 86.4  & 85.4 & 34.6  \\
Real-ESRGAN   & 16.7  & 178.4 & 173 \\
FFDNet        & 0.9   &  14.0  &  2.1  \\
CodeFormer    & 94.1  & 216.3 & 79.8 \\
\midrule
\textbf{GenRes}     & 10.7  & 3803.8 &  1748.2 \\
\textbf{GenRes++ }  & 10.7  & 11411.5 &  4625.6 \\
\midrule
\bottomrule
\end{tabular}
\end{adjustbox}
\end{table}

\begin{table}[t]
\centering
\vspace{-5pt}
\caption{Individual Transform Contribution.}
\vspace{1pt}
\begin{adjustbox}{width=0.6\linewidth}
\begin{tabular}{lcc}
\toprule
\midrule
Transform & ACC (\%) & AP (\%) \\
\midrule
No Transform                       & 91.4          & 96.1          \\
EnlightenGAN                       & 92.2          & 97.3          \\
GFPGAN                             & 91.8          & 96.4          \\
Real-ESRGAN                        & 92.1          & 97.8          \\
FFDNet                             & 92.6          & 97.1          \\
CodeFormer                         & 92.3          & 96.9          \\
\midrule
\textbf{GenRes++}        & \textbf{95.7} & \textbf{99.0} \\
\midrule
\bottomrule
\vspace{-25pt}
\end{tabular}
\label{tab:ablation_individual}
\end{adjustbox}
\end{table}
\vspace{-5pt} \subsection{Qualitative Analysis of Generative Transforms}\vspace{-3pt} 
Figure~\ref{fig:trans} illustrates the effect of each generative transform on representative real and fake images from the BigGAN~\cite{brock2018large} category. Each transform targets a complementary aspect of the image: EnlightenGAN~\cite{jiang2021enlightengan} modifies global luminance and color statistics, GFPGAN~\cite{wang2021gfpgan} introduces a strong GAN-based texture prior, Real-ESRGAN enhances high-frequency components through an upscale–downscale process, FFDNet performs denoising with different responses to real and synthetic noise, and CodeFormer reconstructs local regions via a codebook prior, often leaving a characteristic double-GAN signature in fake images. As shown in the figure~\ref{fig:trans}, these transforms produce visually distinct outputs, indicating that they capture diverse and non-overlapping dimensions of the generative artifact space. Importantly, each transform affects real and fake images differently; for instance, Real-ESRGAN amplifies artifacts in fake images, while FFDNet suppresses natural noise in real images but preserves synthetic patterns. This asymmetric behavior enables learning representations that are robust to appearance changes while remaining sensitive to generative fingerprints, improving generalization to unseen generators.
\vspace{-15pt} 
\subsection{Performance Evaluations and Comparisons}\vspace{-5pt} 
Tables~\ref{tab:detection_acc} and~\ref{tab:detection_ap} present the detection performance in terms of ACC and AP across 19 unseen generative models spanning GAN, other, and diffusion families. The proposed GenRes++ achieves the best overall performance with 95.7\% mACC and 99.1\% mAP, outperforming all prior approaches. Its single-transform variant, GenRes, trained exclusively with FFDNet~\cite{zhang2018ffdnet}, the strongest individual transform, attains a competitive 92.6\% mACC and 97.0\% mAP, demonstrating that even a single well-chosen generative transform yields substantial gains over baselines that do not exploit relational residuals.\\
Across GAN-based models, both GenRes and GenRes++ consistently maintain near-perfect accuracy (above 97\%), with GenRes++ matching or exceeding all baselines including C2P-CLIP~\cite{tan2025c2p} (93.8\%) and FreLens~\cite{chen2025forgelensdataefficientforgeryfocus} (95.0\%). The advantage becomes more pronounced on the other-family methods (\eg, SITD, SAN, CRN, IMLE), where existing methods often struggle. GenRes++ achieves strong results on CRN~\cite{chen2017photographic} (99.4\%) and IMLE~\cite{li2019diverse}(99.5\%), while GenRes, relying solely on FFDNet's denoising residuals, attains 89.5\% and 88.7\% respectively, confirming that denoising-based residuals are informative but benefit substantially from complementary transforms. For diffusion-based models, GenRes++ demonstrates strong robustness across all LDM~\cite{rombach2022high}, and GLIDE~\cite{nichol2021glide} configurations, whereas GenRes shows moderate degradation on Guided Diffusion (71.6\%) and LDM 200 (88.1\%), suggesting that diffusion artifacts are better captured through diverse multi-transform aggregation. In terms of AP, GenRes++ achieves near-perfect scores across nearly all generators, with GenRes also performing strongly (97.0\% mAP), confirming well-separated score distributions between real and fake images even under the single-transform setting. Prior methods tend to specialize in specific generator families, frequency-based approaches perform well on GANs but degrade on diffusion models, while CLIP-based methods improve generalization but remain inconsistent. In contrast, both GenRes and GenRes++ maintain stable performance across all families, with GenRes++ providing the most consistent results. 
\vspace{-4pt}
\subsection{Complexity Analysis}\vspace{-5pt}
Table~\ref{tab:compute_cost} presents the computational cost of the generative transforms used in our pipeline. Among the five models, FFDNet~\cite{zhang2018ffdnet} is the most lightweight, requiring only 0.9M parameters, 14.0 GFLOPs, and 2.1,ms inference time. In contrast, CodeFormer~\cite{zhou2022codeformer} is the most computationally expensive (94.1M parameters, 216.3 GFLOPs, 79.8,ms) due to its transformer-based design. EnlightenGAN~\cite{jiang2021enlightengan} offers a favorable efficiency–performance trade-off with relatively low overhead (8.6M parameters, 16.5 GFLOPs, 5.2,ms), while GFPGAN~\cite{wang2021gfpgan} and Real-ESRGAN~\cite{wang2021realesrgan} lie in the mid-range, with Real-ESRGAN being more compute-intensive despite having fewer parameters than GFPGAN (16.7M vs.\ 86.4M parameters, 178.4 vs.\ 85.4 GFLOPs). In addition, Table~\ref{tab:compute_cost} compares the inference cost of the proposed detection models. Both GenRes and GenRes++ maintain the same number of parameters (10.7M); however, GenRes++ incurs significantly higher computational cost (11411.5 vs.\ 3803.8 GFLOPs) and nearly triples inference time (4625.6 ms vs.\ 1748.2 ms) due to the increased complexity of multi-branch processing. Overall, the detection models exhibit a substantially higher GFLOPs-to-parameter ratio than the generative transforms, highlighting the compute-intensive nature of the detection task.

\subsection{Ablation Study}\vspace{-5pt}
\subsubsection{Individual Transform Contribution}
\vspace{-6pt}
% Table~\ref{tab:ablation_individual} reports the contribution of each generative transformation individually, where a single transform is used alongside the original image during training. Without any transformation, the model achieves an ACC of 91.4\% and AP of 96.1\%, serving as the lower bound. Among individual transforms, FFDNet~\cite{zhang2018ffdnet} (denoising) yields the highest ACC (92.6\%) while Real-ESRGAN~\cite{wang2021realesrgan} (super-resolution) achieves the best AP (97.8\%), suggesting that frequency-domain and up-scaling artifacts are particularly informative cues for detection. Enhancement-based transforms (EnlightenGAN~\cite{jiang2021enlightengan}, CodeFormer~\cite{zhou2022codeformer}, GFPGAN~\cite{wang2021gfpgan}) provide moderate gains, as they alter the generative prior without dramatically changing the frequency signature. Notably, no single transform matches the performance of the full set, GenRes++ combines all five transformations and achieves the best ACC of 95.7\% and AP of 99.0\%, demonstrating that the transforms are complementary and that diversity across enhancement, restoration, super-resolution, and denoising categories is essential for robust cross-generator generalization.
Table~\ref{tab:ablation_individual} evaluates each generative transformation independently, where a single transform is applied alongside the original image during training. Without any transformation, the model achieves 91.4\% ACC and 96.1\% AP. Among individual transforms, FFDNet~\cite{zhang2018ffdnet} yields the highest ACC (92.6\%), while Real-ESRGAN~\cite{wang2021realesrgan} achieves the best AP (97.8\%), indicating that denoising and super-resolution introduce particularly informative cues. Other enhancement-based transforms (EnlightenGAN~\cite{jiang2021enlightengan}, CodeFormer~\cite{zhou2022codeformer}, GFPGAN~\cite{wang2021gfpgan}) provide smaller gains due to more subtle spectral changes. Overall, no single transform matches GenRes++, highlighting that complementary transformations across denoising, enhancement, and super-resolution are critical for robust cross-generator generalization.
\vspace{-12pt}
\subsubsection{Effect of LoRA Rank}
\begin{figure}[!b]
\centering
   \includegraphics[width=0.8\linewidth]{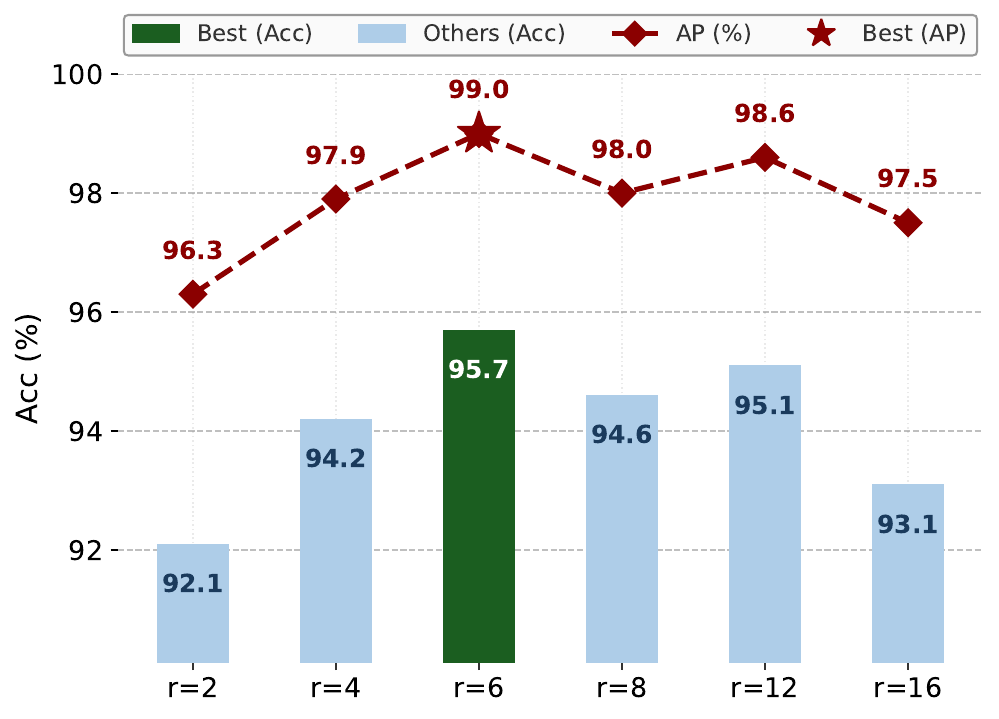}
   \caption{Ablation on LoRA rank $r$. Bars show ACC (\%) and the dashed line shows AP (\%). The best configuration ($r{=}6$) is highlighted in green.}
\label{fig:lora_rank}
\end{figure}

\vspace{-6pt}
% We ablate the LoRA rank $r \in \{2, 4, 6, 12, 16\}$ to analyze the trade-off between adapter capacity and overfitting. As shown in Fig.~\ref{fig:lora_rank}, performance improves from $r{=}2$ (ACC: 92.1\%, AP: 96.3\%) to $r{=}6$ (ACC: 95.7\%, AP: 99.0\%), indicating that a moderate rank is sufficient to effectively adapt the frozen backbone. Beyond $r{=}6$, performance degrades: $r{=}12$ achieves 95.1\% ACC and 98.6\% AP, while $r{=}16$ further drops to 93.1\% ACC and 97.5\% AP. This suggests that higher ranks introduce redundant parameters that lead to overfitting and reduced cross-generator generalization. This trend aligns with prior work on parameter-efficient fine-tuning~\cite{hu2021lora}, which shows that low-rank adapters better preserve the generalization of pretrained representations compared to higher-rank or full fine-tuning approaches. Notably, the performance drop from $r{=}6$ to $r{=}16$ is more pronounced in ACC (2.6\%) than in AP (1.5\%), indicating that accuracy is more sensitive to rank-induced overfitting in this setting. Overall, $r{=}6$ provides the best trade-off between model capacity and generalization.
We ablate the LoRA rank $r \in {2, 4, 6, 8, 12, 16}$ to study the trade-off between adapter capacity and overfitting. As shown in Fig.~\ref{fig:lora_rank}, performance improves from $r{=}2$ (ACC: 92.1\%, AP: 96.3\%) to $r{=}6$ (ACC: 95.7\%, AP: 99.0\%), indicating that a moderate rank is sufficient to adapt the frozen backbone effectively. Beyond this point, performance degrades, with $r{=}12$ and $r{=}16$ dropping to 95.1\%/98.6\% and 93.1\%/97.5\% (ACC/AP), respectively, suggesting over-parameterization harms cross-generator generalization. Overall, $r{=}6$ provides the best balance between model capacity and generalization, consistent with prior findings on low-rank adaptation~\cite{hu2021lora}.
\vspace{-8pt}
\subsubsection{Backbone Selection}\vspace{-5pt}
Table~\ref{tab:backbone} compares three vision backbones under identical training conditions. CLIP ViT-L/14 achieves 91.3\% accuracy and 97.8\% AP, while DINO ViT-L/14 improves slightly to 92.7\% accuracy and 97.6\% AP. Our PE-Core-G14-448 backbone outperforms both by a clear margin, reaching 95.7\% accuracy and 99.0\% AP. The gain stems from PE-Core's larger capacity and its pretraining on diverse visual tasks, which provides richer low-level texture representations that are more sensitive to the subtle statistical inconsistencies introduced by generative models.
\vspace{-6pt}

% \begin{table}[t]
% \centering
% \caption{Backbone comparison. All models use the same NTN head,
% LoRA $r$=6, and five generative transforms.}
% \label{tab:backbone}
% \setlength{\tabcolsep}{8pt}
% \begin{tabular}{lcc}
% \toprule
% Backbone & mAcc.\ (\%) & mAP\ (\%) \\
% \midrule
% CLIP ViT-L/14  & 91.3 & 97.8 \\
% DINO ViT-L/14  & 92.7 & 97.6 \\
% \textbf{PE-Core-G14 (Ours)} & \textbf{95.7} & \textbf{99.0} \\
% \bottomrule
% \end{tabular}
% \end{table}

\begin{table}[t]
\centering
\caption{Backbone Selection.}
\label{tab:backbone}
\vspace{1pt}
\setlength{\tabcolsep}{6pt}
\begin{adjustbox}{width=0.9\columnwidth}
\begin{tabular}{lcc}
\toprule \toprule
Backbone & mACC.\ (\%) & mAP\ (\%) \\
\midrule
CLIP ViT-L/14  & 91.3 & 97.8 \\
DINO ViT-L/14  & 92.7 & 97.6 \\
\midrule
\textbf{PE-Core-G14-448 (Ours)} & \textbf{95.7} & \textbf{99.0} \\
\bottomrule \bottomrule
\end{tabular}
\end{adjustbox}
\vspace{-6pt}
\end{table}

\begin{figure}[t]
\centering
   \includegraphics[width=0.8\linewidth]{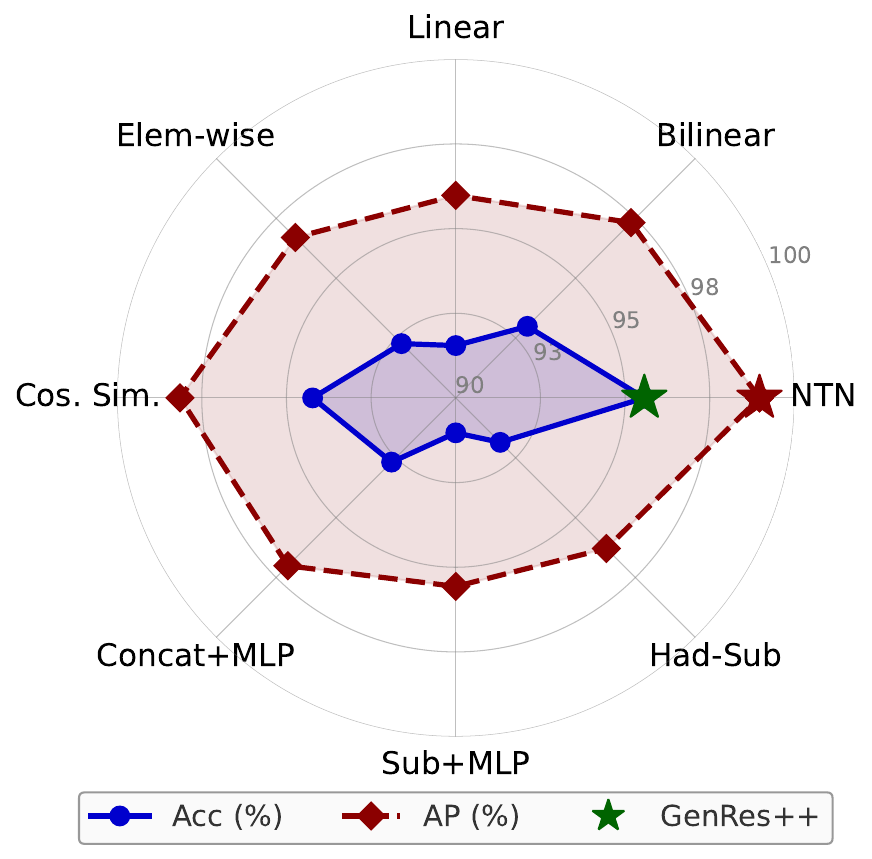}
   \caption{Radar chart comparing fusion functions in the NTN layer. NTN achieves the highest ACC and AP, confirming the effectiveness of bilinear tensor decomposition.}
\label{fig:ab_fusion}
\vspace{-5pt}
\end{figure}
\begin{table}[!t]
\centering
\caption{Leave-One-Out Analysis}
\label{tab:ablation_loo}
\setlength{\tabcolsep}{6pt}
\begin{adjustbox}{width=0.7\columnwidth}
\begin{tabular}{lcc}
\toprule \midrule
Setting & ACC (\%) & AP (\%) \\
\midrule
w/o EnlightenGAN   & 94.1 & 98.2 \\
w/o GFPGAN         & 93.4 & 97.6 \\
w/o Real-ESRGAN    & 93.8 & 97.9 \\
w/o FFDNet         & 92.3 & 96.4 \\
w/o CodeFormer     & 93.1 & 97.3 \\
\midrule
\textbf{GenRes++}  & \textbf{95.7} & \textbf{99.0} \\
\bottomrule \bottomrule
\end{tabular}
\end{adjustbox}
\end{table}
\begin{figure}[t]
\centering
   \includegraphics[width=1\linewidth]{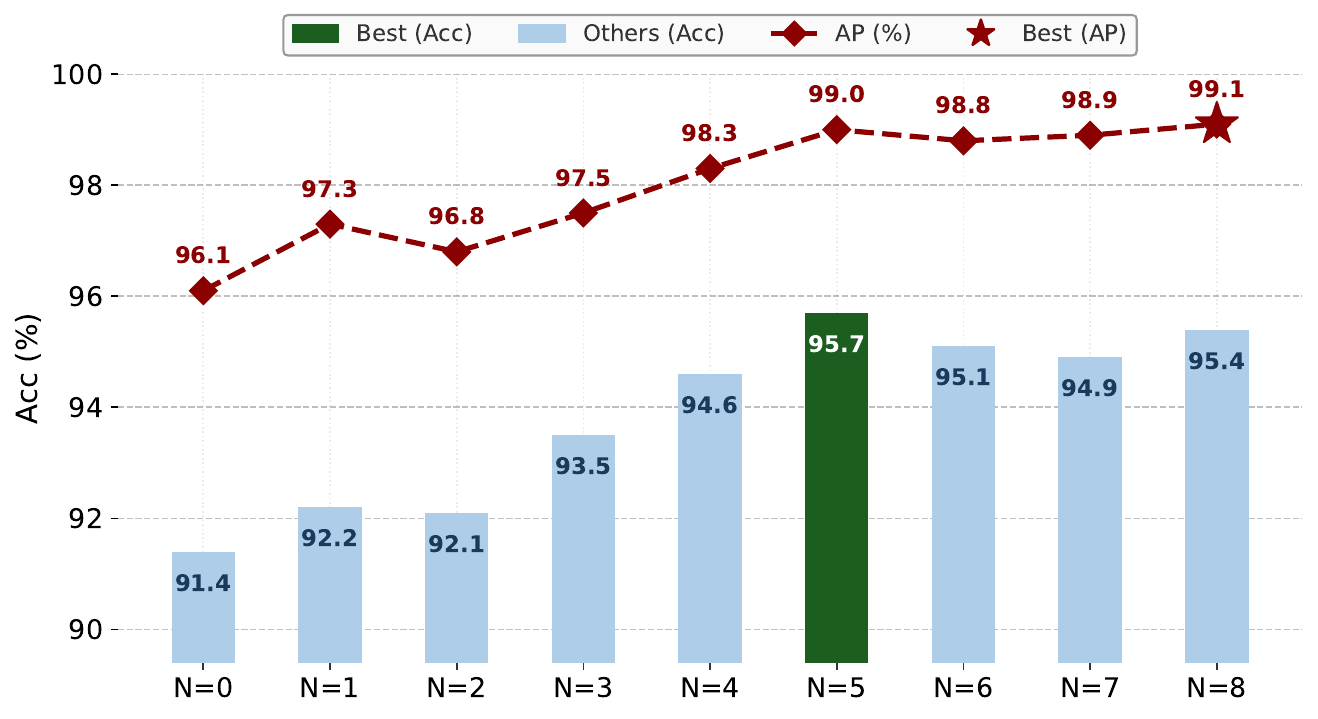}
   \caption{Effect of the number of generative transforms $N$ on detection performance. Bars show ACC (\%), and the dashed line shows AP (\%). Performance peaks at $N{=}5$ 
(green), confirming the selection of five complementary transforms.}
\label{fig:n_trans}
\vspace{-15pt}
\end{figure}
\begin{table}[t]
\centering
\caption{Effect of Perturbations Addition on Performance.}
\vspace{1pt}
\setlength{\tabcolsep}{6pt}
\begin{adjustbox}{width=0.7\columnwidth}
\begin{tabular}{c c c c c}
\toprule
\midrule
Blur & JPEG & Noise & ACC (\%) & AP (\%) \\
\midrule
$\checkmark$ & $\times$ & $\times$ & 86.6 & 94.8 \\
$\times$ & $\checkmark$ & $\times$ & 93.4 & 98.1 \\
$\times$ & $\times$ & $\checkmark$ & 92.7 & 97.5 \\
$\checkmark$ & $\checkmark$ & $\checkmark$ & 84.9 & 93.6 \\
\midrule
\bottomrule
\end{tabular}
\end{adjustbox}
\label{tab:blur_jpeg_noise}
\vspace{-10pt}
\end{table}
\vspace{-5pt}
\subsubsection{Effect of Fusion Function in the NTN Layer}
\vspace{-5pt}
% We compare eight fusion functions within the NTN interaction layer to identify the most effective strategy for combining original and transformed branch features, where $\mathbf{u}$ and $\mathbf{v}$ denote the respective feature vectors. The evaluated methods include NTN, Bilinear (bilinear product), Linear (linear projection), Elem-wise (element-wise Hadamard product), Cos.\ Sim.\ (cosine similarity), Concat+MLP (concatenation followed by an MLP), Sub+MLP (difference vector processed by an MLP), and Had-Sub (Hadamard product minus difference with an MLP). As shown in Fig.~\ref{fig:ab_fusion}, NTN achieves the best performance, reaching 95.7\% ACC and 99.0\% AP, outperforming all alternative fusion strategies. This result highlights the advantage of higher-order feature interactions enabled by NTN, which more effectively capture complementary information between the two branches compared to simpler linear or element-wise operations.
We compare eight fusion strategies within the NTN interaction layer to determine the most effective way to combine original and transformed features, where $\mathbf{u}$ and $\mathbf{v}$ denote the respective embeddings. The evaluated methods include NTN, Bilinear, Linear, element-wise product, cosine similarity, Concat+MLP, Sub+MLP, and Had-Sub fusion. As shown in Fig.~\ref{fig:ab_fusion}, NTN achieves the best performance with 95.7\% ACC and 99.0\% AP, consistently outperforming simpler linear and element-wise alternatives. This demonstrates that higher-order interactions are more effective for capturing complementary relational cues between the two feature branches.
\vspace{-8pt}
\subsubsection{Leave-One-Out Transform Analysis}
\vspace{-5pt}
% To quantify the contribution of each transform, we perform a leave-one-out analysis by removing one transform at a time from the full set of five, while keeping all other components fixed. As given in Table~\ref{tab:ablation_loo}, removing any single transform consistently degrades performance compared to the full configuration (ACC: 95.7\%, AP: 99.0\%), confirming that each transform provides complementary information. The largest drop occurs when removing CodeFormer (ACC: 93.1\%, AP: 97.3\%), indicating that VQ-prior-based face restoration artifacts are highly discriminative. A similar degradation is observed when excluding GFPGAN~\cite{wang2021gfpgan} (ACC: 93.4\%, AP: 97.6\%), highlighting the importance of GAN-based restoration priors. Removing Real-ESRGAN~\cite{wang2021realesrgan} and EnlightenGAN~\cite{jiang2021enlightengan} leads to moderate performance drops, while excluding FFDNet~\cite{zhang2018ffdnet} results in the significant degradation (ACC: 92.3\%, AP: 96.4\%), suggesting that denoising artifacts are partially captured by other transforms. Overall, these results demonstrate that the selected transforms are non-redundant and jointly contribute to robust cross-generator generalization.
To quantify the contribution of each transform, we perform a leave-one-out analysis by removing one transform at a time from the full set of five, while keeping all other components fixed. As shown in Table~\ref{tab:ablation_loo}, removing any transform degrades performance compared to the full model (95.7\% ACC, 99.0\% AP), confirming their complementarity. The largest drop occurs when removing FFDNet (ACC: 92.3\%, AP: 96.4\%), indicating that VQ-prior-based face restoration artifacts are highly discriminative. A similar degradation is observed when excluding GFPGAN~\cite{wang2021gfpgan} (ACC: 93.4\%, AP: 97.6\%), highlighting the importance of GAN-based restoration priors. Removing Real-ESRGAN, EnlightenGAN, and FFDNet also reduces performance, with FFDNet causing notable degradation (92.3\% / 96.4\%). Overall, these results demonstrate that the selected transforms are non-redundant and jointly contribute to robust cross-generator generalization.
\vspace{-8pt}
\subsubsection{Effect of Number of Transforms}
\vspace{-5pt}
We study the impact of the number of generative transforms $N$ used during training on detection performance. Starting from no transformation ($N{=}0$, ACC: 91.4\%, AP: 96.1\%), we progressively add transforms in the order: EnlightenGAN~\cite{jiang2021enlightengan}, GFPGAN~\cite{wang2021gfpgan}, Real-ESRGAN~\cite{wang2021realesrgan}, FFDNet~\cite{zhang2018ffdnet}, CodeFormer~\cite{zhou2022codeformer}, DiffBIR~\cite{lin2024diffbir}, SDEdit~\cite{meng2022sdedit}, and UEGAN~\cite{ni2021uegan}. As shown in Fig.~\ref{fig:n_trans}, performance improves steadily up to $N{=}5$, achieving the best results (ACC: 95.7\%, AP: 99.0\%), indicating that multiple complementary transforms help expose diverse artifact patterns and enhance cross-generator generalization. Beyond $N{=}5$, gains saturate and slight fluctuations are observed: $N{=}6$ yields 95.1\% ACC and 98.8\% AP, $N{=}7$ yields 94.9\% ACC and 98.9\% AP, and $N{=}8$ yields 95.4\% ACC and 99.1\% AP. This suggests that the initial set of transforms, covering enhancement, restoration, super-resolution, and denoising, already captures the major dimensions of artifact diversity. Additional transforms introduce redundancy and may inject conflicting cues, limiting further improvements. Based on these observations, we select $N{=}5$ as the optimal setting, offering the best balance between transform diversity, computational efficiency, and detection performance.
\vspace{-10pt}
\subsubsection{Robustness against Perturbations}
\vspace{-5pt}
Table~\ref{tab:blur_jpeg_noise} evaluates the impact of common perturbations, including Gaussian blur, JPEG compression, and additive noise, on detection performance. For blur, we apply kernels of size $k \in \{3, 5\}$ with $\sigma=0.5$. For JPEG, the quality factor is randomly sampled from $\{70, 75, 80, 85\}$, while Gaussian noise is added with zero mean and standard deviation $\sigma=0.03$. Among individual perturbations, JPEG compression achieves the best performance (93.4\% ACC, 98.1\% AP), followed by Gaussian noise (92.7\% ACC, 97.5\% AP), whereas Gaussian blur performs the worst (86.6\% ACC, 94.8\% AP). This indicates that compression artifacts are more informative for detecting generative inconsistencies than the smoothing effects introduced by blur. Notably, combining all perturbations further degrades performance (84.9\% ACC, 93.6\% AP), suggesting that excessive or conflicting degradations can obscure discriminative cues.
\vspace{-5pt}
\vspace{-5pt}
\section{Limitations and Future Work}
\vspace{-5pt}
Despite its strong cross-generator generalization, GenRes++ has several limitations that suggest directions for future research. First, the framework incurs substantial computational overhead. Each input must be passed through $N{=}5$ separate generative transforms before detection, and the multi-branch design raises the cost of GenRes++ to 11411.5 GFLOPs with an inference time of roughly 4625.6\,ms per image (Table~\ref{tab:compute_cost}), nearly tripling the single-transform GenRes. This cost, dominated by the transform stage and the bilinear NTN interaction, currently limits deployment in real-time or large-scale streaming settings. Future work will explore transform distillation, shared-backbone reuse across branches, and lightweight surrogate residual estimators to approximate the multi-transform signal at a fraction of the cost.\\
Second, the transform set is fixed and manually selected. While the leave-one-out and $N$-ablation studies confirm that the five transforms are complementary and that performance saturates beyond $N{=}5$, the optimal set may shift for new generator families or image domains. Two of the transforms (GFPGAN and CodeFormer) are face-restoration models, which may bias the residual signal toward facial content and reduce its informativeness on non-face imagery. A promising direction is to learn or adaptively select transforms per input rather than relying on a hand-crafted pool.\\
Third, robustness under compound degradation remains limited~\cite{uddin2019anti, uddin2021analysis}. Although the model tolerates individual perturbations reasonably well, accuracy drops to 84.9\% when Gaussian blur, JPEG compression, and additive noise are applied jointly (Table~\ref{tab:blur_jpeg_noise}), indicating that heavy or conflicting post-processing can obscure the generative residual. Improving resilience to real-world laundering (e.g., social-media re-encoding chains) through perturbation-aware or adversarial training is an important next step~\cite{uddin2025guard}.\\
Finally, our evaluation is confined to still images and to the 19 generators of UniversalFakeDetect. The generative-residual principle is, in principle, modality-agnostic, and we plan to extend it to video deepfake detection, where temporal residual consistency offers an additional cue, and to multimodal forgery scenarios that jointly reason over image, audio, and text. Validating the approach on the newest text-to-image and autoregressive generators not represented in the current benchmark is likewise a priority.
\vspace{-8pt}
\section{Conclusion}
\vspace{-5pt}
We presented GenRes and GenRes++, a framework for generalizable AIGI detection grounded in generative residuals, the differential response of real versus synthetic images under secondary generative processing. Relational cues are modeled via a neural tensor network capturing multiplicative cross-feature dependencies, with a Cross-Attention Aggregation module adaptively pooling signals from multiple transforms. We employ a frozen PE-Core ViT with LoRA-based fine-tuning of a large-scale backbone. GenRes++ achieves state-of-the-art 95.7\% mACC and 99.1\% mAP across 19 unseen generative models, outperforming all baselines. Ablation studies confirm that the NTN, multi-transform diversity, LoRA rank $r{=}6$, and $N{=}5$ transforms are each essential to this performance. Future work will extend generative residual learning to video deepfake detection and multimodal forgery scenarios.

{\small
\bibliographystyle{unsrt}
\bibliography{egbib}
}

\end{document}